\documentclass[letterpaper, 10 pt, conference]{ieeeconf}  

\IEEEoverridecommandlockouts                              
\usepackage{microtype}
\usepackage{subfigure}
\usepackage{subcaption}
\usepackage{caption}
\usepackage{booktabs} 
\usepackage{graphicx}
\usepackage{booktabs}
\usepackage{amsmath}
\usepackage{amssymb}
\usepackage{color}
\usepackage{longtable}
\usepackage{etoolbox}
\usepackage[ruled,linesnumbered,lined]{algorithm2e}
\usepackage{algpseudocode} 
\usepackage{hyperref}
\makeatletter
\patchcmd{\@makecaption}
  {\scshape}
  {}
  {}
  {}
\makeatother

\newcommand{\datasetFont}{\text}
\newcommand{\ours}{\datasetFont{UGAT}\xspace}

\title{\LARGE \bf
Uncertainty-aware Grounded Action Transformation \\ towards Sim-to-Real Transfer for Traffic Signal Control
}

\author{Longchao Da, Hao Mei, Romir Sharma and Hua Wei
\thanks{\boldmath Hua Wei with Longchao Da, Hao Mei, are at  
School of Computing and Augmented Intelligence (SCAI), Arizona State University, USA,
        {\tt\small \{hua.wei, longchao, hmei7\}@asu.edu.} Romir Sharma is with the West Windsor-Plainsboro High School South, West Windsor, USA, {\tt\small sharmaromir@gmail.com}}%
\thanks{The work was partially supported by NSF award \#2153311. The views and conclusions contained in this paper are those of the authors and should not be interpreted as representing any funding agencies.}
}

\begin{document}

\maketitle
\thispagestyle{empty}
\pagestyle{empty}

\begin{abstract}

Traffic signal control (TSC) is a complex and important task that affects the daily lives of millions of people. Reinforcement Learning (RL) has shown promising results in optimizing traffic signal control, but current RL-based TSC methods are mainly trained in simulation and suffer from the performance gap between simulation and the real world. In this paper, we propose a simulation-to-real-world (sim-to-real) transfer approach called \ours, which transfers a learned policy trained from a simulated environment to a real-world environment by dynamically transforming actions in the simulation with uncertainty to mitigate the domain gap of transition dynamics. We evaluate our method on a simulated traffic environment and show that it significantly improves the performance of the transferred RL policy in the real world. 

\end{abstract}

\section{Introduction}
Traffic Signal Control (TSC) is vital for enhancing traffic flow, alleviating congestion in contemporary transportation systems, and providing widespread societal benefits. It remains an active research area due to the intricate nature of the problem. TSC must cope with dynamic traffic scenarios, necessitating the development of adaptable algorithms to respond to changing conditions.

Recent advances in reinforcement learning (RL) techniques have shown superiority over traditional approaches in TSC~\cite{noaeen2022reinforcement}. In RL, an agent aims to learn a policy through trial and error by interacting with an environment to maximize the cumulative expected reward over time. The biggest advantage of RL is that it can directly learn how to generate adaptive signal plans by observing the feedback from the environment.

\begin{figure*}[htbp]
    \centering
    \includegraphics[width=0.95\textwidth]{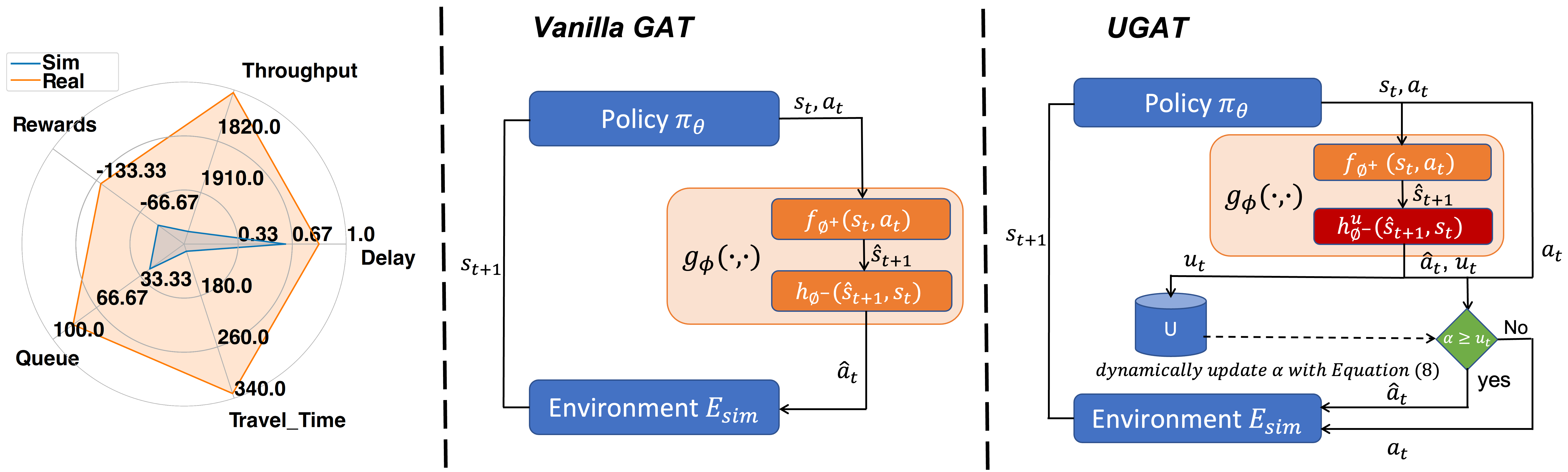}
    \caption{The performance gap in sim-to-real transfer and the schematic of GAT and \ours.
    \textbf{Left}: The method~\cite{wei2018intellilight} trained in simulation has a performance drop when transferred to the real world in all five evaluation metrics in TSC.
    \textbf{Middle}: GAT method takes grounded action $\hat{a}_t$ when a policy returns an action $a_t$ from the $E_{sim}$. Grounded actions taken with high model uncertainty on $g_{\phi(\cdot, \cdot)}$ will enlarge the transition between $P_{\theta}$ and $P^*$, making the gap between $E_{sim}$ and $E_{real}$ large and policy learning step not stable.
    \textbf{Right}: \ours quantifies the model's uncertainty and decide to take or reject the grounded action $\hat{a}_t$ based on the current output of model uncertainty $u_t$ given the current $s_t$ state and action $a_t$. This behavior will mitigate the gap between $P_{\phi}$ and $P^*$ and make the policy learning step stable.}
    \label{fig:intro}
    \vspace{-5mm}
\end{figure*}

One major issue of applying current RL-based TSC approaches in the real world is that these methods are mostly trained in simulation and suffer from the performance gap between simulation and the real world. 
While training in simulations offers a cost-effective means to develop RL-based policies, it may not fully capture the complexities of real-world dynamics, limiting RL-based TSC models' practical performance~\cite{jiang2021simgan}. Simulators often employ static vehicle settings, such as default acceleration and deceleration, whereas real-world conditions introduce substantial variability influenced by factors like weather and vehicle types. These inherent disparities between simulation and reality impede RL-based models trained in simulations from achieving comparable real-world performance, as depicted in Figure~\ref{fig:intro}.

To bridge this gap, prior research has concentrated on enhancing traffic simulators to align more closely with real-world conditions, using real-world data~\cite{zhang2019cityflow}. This enables smoother policy or model transfer from simulation to reality, minimizing performance disparities. Yet, altering internal simulator parameters can be challenging in practice. To tackle this issue, Grounded Action Transformation (GAT) has emerged as a popular technique, aiming to align simulator transitions more closely with reality. However, GAT has predominantly been applied to robotics, with limited exploration in the context of traffic signal control.

In this paper, we present Uncertainty-aware Grounded Action Transformation (\ours), an approach that bridges the domain gap of transition dynamics by dynamically transforming actions in the simulation with uncertainty. 
\ours learns to mitigate the discrepancy between the simulated and real-world dynamics under the framework of grounded action transformation (GAT), which learns an inverse model that can generate an action to ground the next state in the real world with a desired next state predicted by the forward model learned in simulation. Specifically, to avoid enlarging the transition dynamics gap induced by the grounding actions with high uncertainty, \ours dynamically decides when to transform the actions by quantifying the uncertainty in the forward model. Our experiments demonstrate the existence of the performance gap in traffic signal control problems and further show that \ours has a good performance in mitigating the gap with higher efficiency and stability.

\section{Preliminaries}


This section will formalize the traffic signal control (TSC) problem and its RL solutions and introduce the grounded action transformation (GAT) framework for sim-to-real transfer.

\subsection{Concepts of TSC and  RL Solutions}

    
In the TSC problem, each traffic signal controller decides the phase of an intersection, which is a set of pre-defined combinations of traffic movements that do not conflict while passing through the intersection. Given the current condition of this intersection, the traffic signal controller will choose a phase for the next time interval $\Delta t$ to minimize the average queue length on lanes around this intersection. Following existing work~\cite{wei2018intellilight}, an agent is assigned to each traffic signal, and the agent will choose actions to decide the phase in the next $\Delta t$. The TSC problem is defined as an MDP which could be characterized by $\mathcal{M}  = \langle \mathcal{S}, \mathcal{A}, P, r, \gamma \rangle$ where the state space $\mathcal{S}$ contains each lane's number of vehicles and the current phase of the intersection in the form of the one-hot code, $s_t \in \mathcal{S}$. Action space (discrete) $\mathcal{A}$ is the phase chosen for the next time interval $\Delta t$. Transition dynamics $P(s_{t+1}| s_t, a_t)$ maps $\mathcal{S} \times \mathcal{A} \rightarrow \mathcal{S}$, describing the probability distribution of next state $s_{t+1} \in \mathcal{S}$. Reward $r$ is an immediate scalar return from the environment calculated as $r_t = -\sum_l w_t^l$, where $l$ is the lane belonging to the intersection and $w_t^l$ is the queue length at each lane. And the $\gamma$ is the discount factor. Policy $\pi_{\theta}$ could be represented as the logic of: $\mathcal{S} \rightarrow \mathcal{A}$. An RL approach solves this problem by maximizing the long-term expectation of discounted accumulation reward adjusted $\gamma$. The discounted accumulated reward is $\mathbb{E}_{(s_t, a_t)\sim (\pi_{\theta},\mathcal{M})}[ \sum_{t=0}^T \gamma^{T-t} r_t(s_t, a_t) ]$. Since the action space $\mathcal{A}$ is discrete, we follow the past work using Deep Q-network (DQN)~\cite{wei2018intellilight} to optimize the $\pi_{\theta}$, the above procedure is conducted in simulation environment $E_{sim}$.

\subsection{Grounded Action Transformation}


Grounded action transformation (GAT) is a framework originally proposed in robotics to improve robotic learning by using trajectories from the physical world $E_{real}$ to modify $E_{sim}$. Under the GAT framework, MDP in $E_{sim}$ is imperfect and modifiable, and it can be parameterized as a transition dynamic $P_{\phi}(\cdot|s, a)$. Given real-world dataset $\mathcal{D}_{real} =  \{\tau^{1}, \tau^{2}, \dots, \tau^{I}\}$, where $\tau^{i} = (s_0^{i}, a_0^{i}, s_1^{i}, a_1^{i}, \dots, s_{T-1}^{i}, a_{T-1}^{i}, s_T^{i})$ is a trajectory collected by running a policy $\pi_{\theta}$ in $E_{real}$, GAT aims to minimize differences between transition dynamics by finding $\phi^*$:
\vspace{-2mm}
\begin{equation}
\phi^* = \arg \min_{\phi} \sum_{\tau^i \in \mathcal{D}_{real}} \sum_{t=0}^{T-1} d(P^*(s^i_{t+1}|s^i_t, a^i_t), P_{\phi}(s^i_{t+1}|s^i_t, a^i_t))
\end{equation}


\noindent where $d(\cdot)$ is the distance between two dynamics, $P^*$ is the real world transition dynamics, and $P_{\phi}$ is the simulation transition dynamics. 

To find $\phi$ efficiently, GAT takes the agent's state $s_t$ and action $a_t$ predicted by policy $\pi_\theta$ as input and outputs a grounded action $\hat{a}_t$. Specifically, it uses an action transformation function parameterized with $\phi$: 
\vspace{-2mm}
\begin{equation}
\label{eq:gat}
    \hat{a}_t = g_{\phi}(s_t, a_t) = h_{\phi^{-}}(s_t, f_{\phi^{+}}(s_t, a_t))
    \vspace{-2mm}
\end{equation}
 which includes two specific functions: a forward model $f_{\phi^{+}}$, and an inverse model $h_{\phi^{-}}$, as is shown in Fig.~\ref{fig:intro}.   
 
\noindent $\bullet$ \textit{The forward model} $f_{\phi^{+}}$ is trained with the data from $E_{real}$, aiming to predict the next possible state $\hat{s}_{t+1}$ given current state $s_t$ and action $a_t$:
\vspace{-2mm}
\begin{equation}
\label{eq:forward}
     \hat{s}_{t+1} =  f_{\phi^{+}}(s_t, a_t)
     \vspace{-2mm}
\end{equation}

\noindent $\bullet$ \textit{The inverse model} $h_{\phi^{-}}$ is trained with the data from $E_{sim}$, aiming to predict the possible action $\hat{a}_t$ that could lead the current state $s_t$ to the given next state. Specifically, the inverse model in GAT takes $\hat{s}_{t+1}$, the output from the forward model, as its input for the next state: 
\vspace{-2mm}
\begin{equation}
\label{eq:inverse}
     \hat{a}_t =  h_{\phi^{-}}(\hat{s}_{t+1}, s_t)
     \vspace{-2mm}
\end{equation}



Given current state $s_t$ and the action $a_t$ predicted by the policy $\pi_\theta$, the grounded action $\hat{a}_t$ takes place in $E_{sim}$ will make the resulted $s_{t+1}$ in $E_{sim}$ close to the predicted next state $\hat{s}_{t+1}$ in $E_{real}$, which makes the dynamics $P_{\phi}(s_{t+1}|s_t, \hat{a}_t)$ in simulation close to the real-world dynamics $P^*(\hat{s}_{t+1}|s_t, a_t)$. Therefore, the policy $\pi_\theta$ is learned in $E_{sim}$ with $P_{\phi}$ close to $P^*$ will have a smaller performance gap when transferred to $E_{real}$ with $P^*$.





\vspace{-4mm}
\section{Methods}


To mitigate the gap in the transition dynamics between traffic simulations and real-world traffic systems, we use the vanilla GAT and analyze its limitations. To overcome it's problem, we 
propose \ours to further leverage uncertainty quantification to take grounded action dynamically. 

\subsection{Vanilla GAT for TSC}

We use the vanilla GAT for the traffic signal control problem by specifying the learning of $f_{\phi^+}$ and $h_{\phi^-}$: 

\noindent $\bullet$ \textit{The forward model} $f_{\phi^+}(s_t, a_t)$ in traffic signal control problem predicts the next traffic state $\hat{s}_{t+1}$ in the real world given taken action $a_t$ and the current traffic state $s_t$. We approximate $f_{\phi^+}$ with a deep neural network and optimize $\phi^+$ by minimizing the Mean Squared Error (MSE) loss:
\begin{equation}
\label{eq:forward-loss}
     \mathcal{L}(\phi^+) = MSE(\hat{s}^i_{t+1}, s^i_{t+1}) = MSE(f_{\phi^+}(s^i_t, a^i_t), s^i_{t+1})
\end{equation}
where $s^i_t$, $a^i_t$, $s^i_{t+1}$ are sampled from the trajectories collected from $E_{real}$.

\noindent $\bullet$ \textit{The inverse model} $h_{\phi^-}(\hat{s}_{t+1}, s_t)$ in traffic signal control predicts the grounded action $\hat{a}^i_t$ in simulation $E_{sim}$ to reproduce the same traffic states $\hat{s}_{t+1}$. 
We approximate $h_{\phi^-}$ with a deep neural network and optimize $\phi^-$ by minimizing the Categorical Cross-Entropy (CCE) loss since the target $a^i_t$ is a discrete value:
\vspace{-2mm}
\begin{equation}
\label{eq:inverse-loss}
    \mathcal{L}(\phi^-) = CCE(\hat{a}^i_t, a^i_t) = CCE(h_{\phi^-}(s^i_{t+1}, s^i_t), a^i_t)
\vspace{-2mm}
\end{equation}
where $s^i_t$, $a^i_t$, $s^i_{t+1}$ are sampled from the $E_{sim}$ trajectories.

\vspace{-1mm}
\subsection{Uncertainty-aware GAT}
In this section, we will introduce the limitations of the vanilla GAT and propose an uncertainty-aware method on GAT that can benefit from quantifying model uncertainty.


\subsubsection{\textbf{Model Uncertainty on $g_{\phi}$}}


The vanilla GAT takes supervised learning to train the action transformation function $g_{\phi}$, and grounded action transformation $\hat{a}$ is taken at each step while improving in the $E_{sim}$. However, the action transformation function $g_{\phi}$ could have high model uncertainty on unseen state and action inputs, which is likely to happen during the exploration of RL. With high model uncertainty on $g_{\phi}$, the grounded action $\hat{a}$ in Equation~\eqref{eq:gat} is likely to enlarge the performance gap instead of mitigating it if the high uncertainty action is taken because it will make policy learning unstable and hard to converge.



To overcome the enlarged gap induced by $\hat{a}$ with high model uncertainty in $g_{\phi}$, we need uncertainty quantification methods~\cite{kabir2018neural} to keep track of the uncertainty of $g_{\phi}$. Specifically, we would like the action transformation function to output an uncertainty value $u_t$ in addition to $\hat{a}_t$:
\vspace{-2mm}
\begin{equation}
\label{eq:ugat-uncertainty}
    \hat{a}_t, u_t = g_{\phi}(s_t, a_t) = h_{\phi^-}(f_{\phi^+}(s_t, a_t), s_t)
\vspace{-2mm}
\end{equation}

In general, any methods capable of quantifying the uncertainty of a predicted class from a deep neural network (since $h_{\phi^-}$ is implemented with deep neural networks) could be utilized, like evidential deep learning (EDL), Concrete Dropout~\cite{gal2017concrete}, Deep Ensembles~\cite{lakshminarayanan2017simple}, etc. In this paper, we explored different state-of-the-art uncertainty quantification methods and found out that they all perform well with our method (their experimental results can be found in Section~\ref{sec:exp:uncertainty}). We adopted EDL as the default in our method as it performs the best with our method.

Intuitively, during action grounding, whenever model $g_{\phi}(s_t, a_t)$ returns a grounded action $\hat{a}_t$, if the uncertainty $u_t$ is less than the threshold $\alpha$, the grounded action $\hat{a}_t$ will be taken in the simulation environment $E_{sim}$ for policy improvement; otherwise, we will reject $\hat{a}_t$ and take the original $a_t$. This uncertainty quantification allows us to evaluate the reliability of the transformation model and take grounded actions $\hat{a}$ when the model is certain that the resulting transition $P_{\phi}(s_t, \hat{a}_t)$ would mirror that of the real-world environment $E_{real}$ transition $P^*(s_t, a_t)$. This process enables us to minimize the gap in Equation~\eqref{eq:gat} between the policy training environment $E_{sim}$ and the policy testing environment $E_{real}$, thereby mitigating the performance gap.

\subsubsection{\textbf{Dynamic Grounding Rate $\alpha$}}
The threshold $\alpha$, which we referred to as the grounding rate,  helps us to decide when to filter out $\hat{a}_t$ with uncertainty $u_t$. One naive approach of deciding the grounding rate $\alpha$ is to treat it as a hyperparameter for training and keep it fixed during the training process. However, since $g_{\phi}(s_t, a_t)$ keeps being updated during the training process, the model uncertainty of $g_{\phi}$ is dynamically changing. Even with the same $s_t$ and $a_t$, the output $u_t$ and $\hat{a}_t$ from $g_{\phi}(s_t, a_t)$ could be different in different training iterations. 

An alternative yet feasible approach is to set grounding rate $\alpha$ dynamically changing with the model uncertainty during different training iterations.
To dynamically adjust the grounding rate with the changing of model uncertainty, we keep track of the model uncertainty $u_t$ of $g_{\phi}(s_t, a_t)$ during each training iteration. 
At the end of each iteration $i$, we update the grounding rate $\alpha$ for the next iteration based on the past record of model uncertainty by calculating the mean
\vspace{-2mm}
\begin{equation}
\label{eq:u-update}
    \alpha = \frac{\sum^E_{e=1} \sum^{T-1}_{t=0} u^e_t}{T \times E}
\vspace{-2mm}
\end{equation}
from the logged uncertainties in the last $E$ epochs.
This dynamic grounding rate $\alpha$ can synchronously adjust $\alpha$ with the update of $g_{\phi}$ and relief efforts on hyper-parameter tuning.

\vspace{-3mm}
\subsection{Training Algorithm}

The overall algorithm for \ours is shown in Algorithm~\ref{algo:UGAT}.
We begin by pre-training the RL policy $\pi_{\theta}$ for $M$ epochs in the simulation environment $E_{sim}$. In each training iteration of \ours, we collect datasets for both $E_{sim}$ and $E_{real}$, following the data collection process from~\cite{hanna2017grounded}. It's worth noting that data collection in $E_{real}$ does not necessarily occur during the training process; it can be obtained from existing offline logged data. With the collected data, we update $g_{\phi}$ by training both the forward model $f_{\phi^+}$ and the inverse model $h_{\phi^-}$. Using the updated $g_{\phi}$, we employ policy $\pi_\theta$ to interact with $E_{sim}$ for further policy training. Prior to executing the action $a_t$ generated by $\pi_\theta(s_t)$ in the environment $E_{sim}$, \ours grounds the actions using $\hat{a}t$ and the model uncertainty $u_t$ from $g{\phi}(s_t, a_t)$. If the model uncertainty $u_t$ surpasses the grounding rate $\alpha$, the grounded action $\hat{a}t$ is rejected, and the original action $a_t$ is executed in the simulation $E{sim}$. Subsequently, $u_t$ is added to the logged uncertainty $U$. The RL policy $\pi_{\theta}$ undergoes updates during the interaction with $E_{sim}$. After $E$ rounds of intersections, we update $\alpha$ using Equation~\eqref{eq:u-update} to prepare for the next round of policy training.

\begin{algorithm}[h!]
\DontPrintSemicolon
\caption{Algorithm for \ours with model uncertainty quantification}
\label{algo:UGAT}
\KwIn{Initial policy $\pi_{\theta}$, forward model $f_{\phi^+}$, inverse model $h_{\phi^-}$, real-world dataset $\mathcal{D}_{real}$, simulation dataset $\mathcal{D}_{sim}$, grounding rate $\alpha = \inf$}
\KwOut{Policy $\pi_{\theta}$, $f_{\phi^+}$, $h_{\phi^-}$}

    Pre-train policy $\pi_{\theta}$ for M iterations in $E_{sim}$ \;
    
	\For {i = 1,2, \dots, I}
	{
        Rollout policy $\pi_{\theta}$ in $E_{sim}$ and add data to $\mathcal{D}_{sim}$ \;
        
        Rollout policy $\pi_{\theta}$ in $E_{real}$ and add data to $\mathcal{D}_{real}$ \\

        \# \textbf{\textit{Transformation function update step}} \;
        
        Update $f_{\phi^+}$ with Equation~\eqref{eq:forward-loss} \;
        Update $h_{\phi^-}$ with Equation~\eqref{eq:inverse-loss}  \;
        
        Reset logged uncertainty $U^i = List()$ \;
        
        \# \textbf{\textit{Policy training}}\;
        \For {e = 1, 2, \dots, E}
        {
        \# \textbf{\textit{Action grounding step}} \;
            \For {t = 0, 1 ,\dots, T-1}
            {
            $a_t = \pi(s_t)$ \;
            Calculate $\hat{a}_t$ and $u_t$ with Equation~\eqref{eq:ugat-uncertainty} \;
            \If{$u^e_t \geq \alpha $ }   
                {
                
                $\hat{a}_t = a_t$  \textbf{\textit{\# Reject grounded action}}
                }
            $U.append(u^e_t)$ \;
            }
            \# \textbf{\textit{Policy update step}}\;
            Improve policy $\pi_{\theta}$ with reinforcement learning\;
        }  
       Update $\alpha$ with Equation~\eqref{eq:u-update} \;
    }
    \vspace{-5mm}
\end{algorithm} 

\section{Experiment and Results}


In this section, we investigate several aspects of our study: the presence of a performance gap in TSC, the effectiveness of \ours in mitigating this gap, the influence of dynamic grounding rate $\alpha$, uncertainty quantification, and action grounding on \ours' performance, and the stability of \ours across various uncertainty quantification methods.

\vspace{-2mm}
\subsection{\textbf{Experiment Settings}}
In this section, we introduce the overall environment setup for our experiments, and commonly used metrics. \\

\begin{table}[htb]
\vspace{-5mm}
\small
\centering
\caption{Real-world Configurations for $E_{real}$}
\label{param}
\setlength{\tabcolsep}{1mm}

\begin{tabular}{cccccc}
\toprule
Setting & \begin{tabular}[c]{@{}c@{}}accel \\ (m/$s^2$)\end{tabular} & \begin{tabular}[c]{@{}c@{}}decel \\ (m/$s^2$)\end{tabular} & \begin{tabular}[c]{@{}c@{}}eDecel \\ (m/$s^2$)\end{tabular} & \begin{tabular}[c]{@{}c@{}}sDelay \\ (s)\end{tabular} & Description        \\ 
\midrule
Default & 2.60 & 4.50 & 9.00 & 0.00 & --- \\
V1   &   1.00     &  2.50 & 6.00 & 0.50 & Lighter loaded vehicles   \\
V2   &   1.00     &  2.50 & 6.00 & 0.75 & Heavier loaded vehicles \\
V3   &   \textcolor{black}{0.75}     &  \textcolor{black}{3.50} & \textcolor{black}{6.00} & \textcolor{black}{0.25}    &  Rainy weather\\
V4   &   \textcolor{black}{0.50}     &  \textcolor{black}{1.50} & \textcolor{black}{2.00} & \textcolor{black}{0.50} & Snowy weather   \\
\bottomrule
\end{tabular}
\end{table}

\subsubsection{\textbf{Environment Setup}}  
In this paper, we implement \ours upon LibSignal~\cite{mei2022libsignal}, an open-sourced traffic signal control library that integrates multiple simulation environments. We treat Cityflow \cite{zhang2019cityflow} as the simulation environment $E_{sim}$ and SUMO \cite{lopez2018microscopic} as the real-world environment $E_{real}$. In later sections, we use $E_{sim}$ and $E_{real}$ by default unless specified. To mimic real-world settings, we consider four configurations in SUMO under two types of real-world scenarios: heavy industry roads and special weather-conditioned roads, with their specific parameters defined in Table~\ref{param}. 
\\
$\bullet$ {\emph{Default setting~\footnote{\url{https://sumo.dlr.de/docs/Definition_of_Vehicles,_Vehicle_Types,_and_Routes.html}}}}. This is the default parameters for SUMO and CityFlow which describe the normal settings of the vehicle's movement in $E_{sim}$, with 8 phases TSC strategy.
\\
$\bullet$ {\emph{Heavy industry roads.}}
We model the places where the majority of vehicles could be heavy trucks. In Table \ref{param}, for the vehicles in $V1$ and $V2$, their accelerating, decelerating, and emergency decelerating rates are more likely to be slower than the default settings. We further consider the vehicles' average startup delay (larger than the default assumption $0s$). As shown in Table \ref{param}, $V1$ describes roads with lighter-loaded vehicles while $V2$ describes the same roads with heavier-loaded vehicles, and they vary at startup delay.
\\
$\bullet$ {\emph{Special weather-conditioned roads.}}
We examine scenarios with adverse weather conditions, specifically rainy ($V3$) and snowy ($V4$) conditions, as outlined in Table \ref{param}. In these settings, vehicle acceleration, deceleration, and emergency deceleration rates are reduced compared to the default values, while startup delays are increased. Notably, in snowy weather, the first three rates are lower than in rainy conditions, and the startup delay difference is extended to simulate tire slip.

\subsubsection{\textbf{Evaluation Metrics}} 
Following the literatures in TSC~\cite{ wei2021recent}, we adopt commonly used traffic signal control metrics as below: \\
$\bullet$ \textit{Average Travel Time (ATT)} is the average time $t$ it takes for a vehicle to travel through a specific section of a road network. For a control policy, the smaller $ATT$, the better. \\
$\bullet$ \textit{Throughput (TP)} is the number of vehicles reached their destinations given amount of time. The larger $TP$, the better.\\
$\bullet$ \textit{Reward} is an RL term that measures the return by taking action $a_t$ under state $s_t$. We use the total number of waiting vehicles as the reward, aligned with Preliminaries. The larger the reward, the fewer waiting vehicles, the better.\\
$\bullet$ \textit{Queue} is the number of vehicles waiting to pass through a certain intersection in the road network. Smaller is better.\\
$\bullet$ \textit{Delay} is the average delay per vehicle in seconds and measures the amount of time that a vehicle spends waiting in the network. Smaller is better.

\begin{table*}[htb]
\small
\caption{The performance using \textbf{Direct-Transfer} method compared with using \textbf{\ours} method. The ($\cdot$) shows the metric gap $\psi_{\Delta}$ from $E_{real}$ to $E_{sim}$ and the $\pm$ shows the standard deviation with 3 runs. The $\uparrow$ means that the higher value for the metric indicates a better performance and $\downarrow$ means that the lower value indicates a better performance.}
\label{tab:result}

\centering
\setlength{\tabcolsep}{1mm}
\scalebox{0.76}{
\begin{tabular}{c|ccccc|ccccc}
\toprule
Setting & \multicolumn{5}{c}{Direct Transfer} & \multicolumn{5}{c}{UGAT} \\ 

\cmidrule(lr){2-6}\cmidrule(lr){7-11}


&$ATT(\Delta\downarrow)$ &$TP(\Delta\uparrow)$ &$Reward(\Delta\uparrow)$ &$Queue(\Delta\downarrow)$ &$Delay(\Delta\downarrow)$ &$ATT(\Delta\downarrow)$  &$TP(\Delta\uparrow)$  &$Reward(\Delta\uparrow)$ &$Queue(\Delta\downarrow)$ &$Delay(\Delta\downarrow)$ \\

\midrule

V1    
     & 158.93(47.69) & 1901(-77) & -71.55(-32.11) & 47.71(21.59) & 0.73(0.11)

     & \textbf{144.72(33.49)$_{\pm{\text{3.61}}}$} & \textbf{1925(-52)$_{\pm{\text{4.58}}}$} & \textbf{-59.38(-19.94)$_{\pm{\text{3.08}}}$} & \textbf{39.58(13.47)$_{\pm{\text{2.04}}}$} & \textbf{0.67(0.05)$_{\pm{\text{0.01}}}$} \\
     
V2  
     & 177.27(66.03) & 1898(-80) & -87.71(-48.27) & 
     58.59(32.47) & 0.76(0.14) 

     & \textbf{164.65(53.52)$_{\pm{\text{12.94}}}$} 
     & \textbf{1907(-71)$_{\pm{\text{13.06}}}$} 
     & \textbf{-75.18(-35.74}$_{\pm{\text{8.37}}}$ 
     & \textbf{50.25(24.14)}$_{\pm{\text{5.56}}}$ 
     & \textbf{0.72(0.10)}$_{\pm{\text{0.01}}}$ \\

V3   
     &  205.86(94.63) & 1877(-101) & -101.26(-61.82) & 67.62(41.51) & 0.76(0.14)      & 
     
     \textbf{183.22(71.99)$_{\pm{\text{13.22}}}$} & \textbf{1900(-78)$_{\pm{\text{13.08}}}$} & \textbf{-82.38(-42.94)$_{\pm{\text{9.11}}}$} & \textbf{55.05(28.94)$_{\pm{\text{6.08}}}$} & \textbf{0.72(0.10)$_{\pm{\text{0.01}}}$} \\

V4   
    & 332.48(221.25) & 1735(-252) & -126.71(-87.23) & 84.53(58.42) & 0.83(0.21) &      
    
    \textbf{284.26(173.03)$_{\pm{\text{6.67}}}$} & \textbf{1794(-184)$_{\pm{\text{12.05}}}$} & \textbf{-111.68(-72.24)$_{\pm{\text{7.25}}}$} & \textbf{74.54(48.43)$_{\pm{\text{4.82}}}$} & \textbf{0.8(0.18)$_{\pm{\text{0.01}}}$} \\

\bottomrule
\end{tabular}
}
\label{tab:main}
\vspace{-4mm}
\end{table*}

In this work, our goal is to mitigate the performance gap of trained policy $\pi_{\theta}$  between $E_{sim}$ and $E_{real}$, we additionally calculate the gap $\Delta$ for each referred metric as $ATT_{\Delta}$, $TP_{\Delta}$, $Reward_{\Delta}$, $Queue_{\Delta}$, and $Delay_{\Delta}$. 
For certain metric $\psi$:
\begin{equation} \label{eq:delta}
    \psi_{\Delta} = \psi_{real} - \psi_{sim}
    \vspace{-2mm}
\end{equation}

Because in real-world settings, policy $\pi_{\theta}$ tends to perform worse than in simulation, so the $ATT$, $Queue$, and $Delay$ in $E_{real}$ are normally larger than those in $E_{sim}$. Based on the goal of mitigating the gap, improving $\pi_{\theta}$ performance in  $E_{sim}$, we expect: for $ATT_{\Delta}$, $Queue_{\Delta}$, and $Delay_{\Delta}$, the smaller the better. Because $TP_{\Delta}$, $Reward_{\Delta}$ will be negative values, the larger, the better.

\subsection{\textbf{Experiment Results}}

\subsubsection{\textbf{Gap between real-world and simulator}}
\begin{figure}[htb]
    \centering
    \begin{tabular}{cc}
    \includegraphics[width=0.220\textwidth]     
    {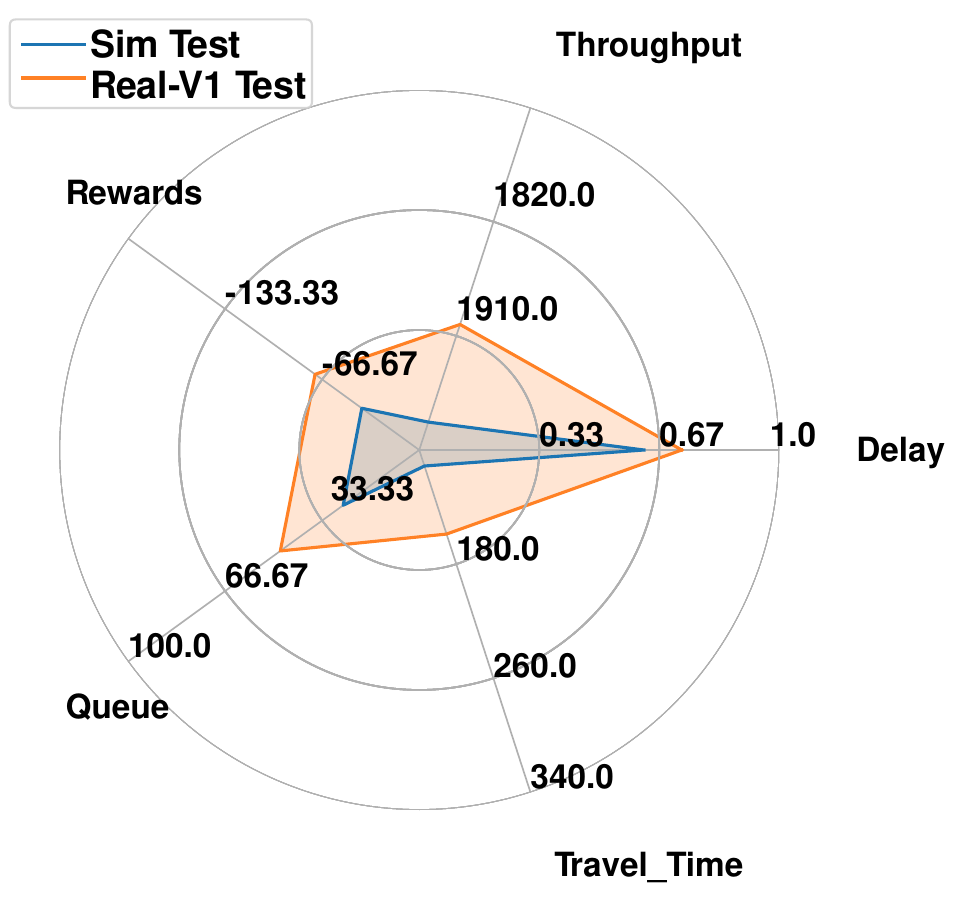}
    &\includegraphics[width=0.220\textwidth]
    {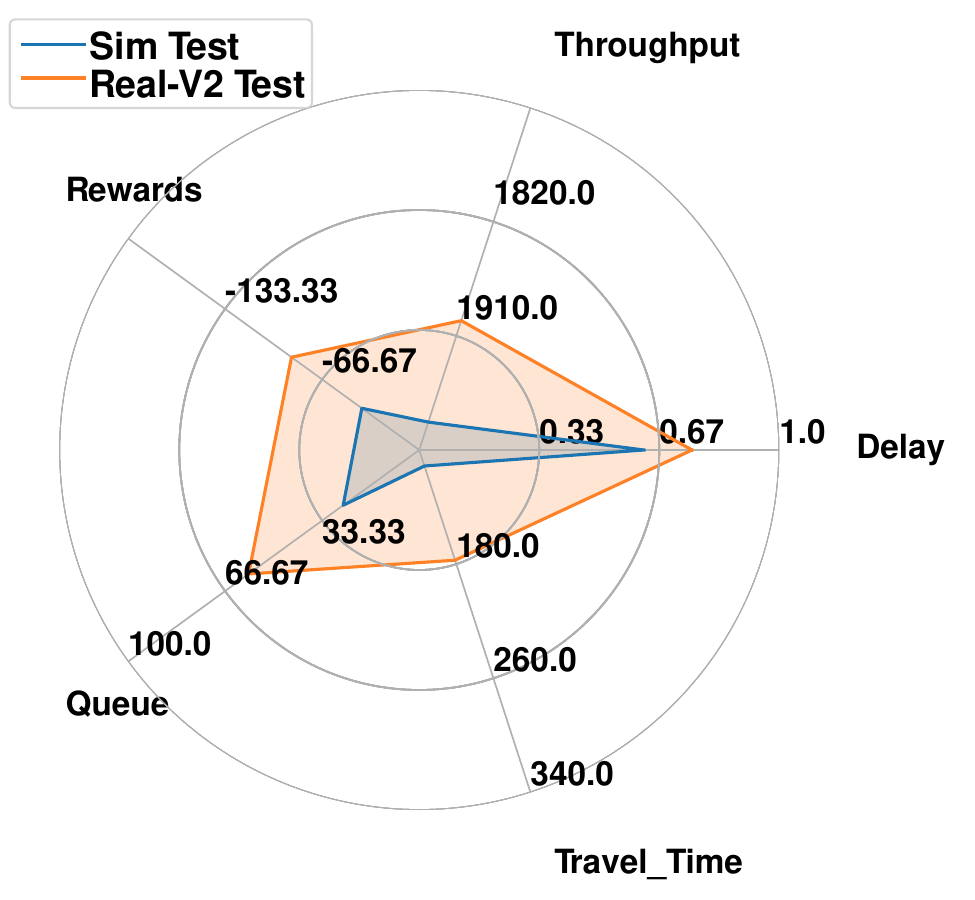}\\
    \includegraphics[width=0.220\textwidth]     
    {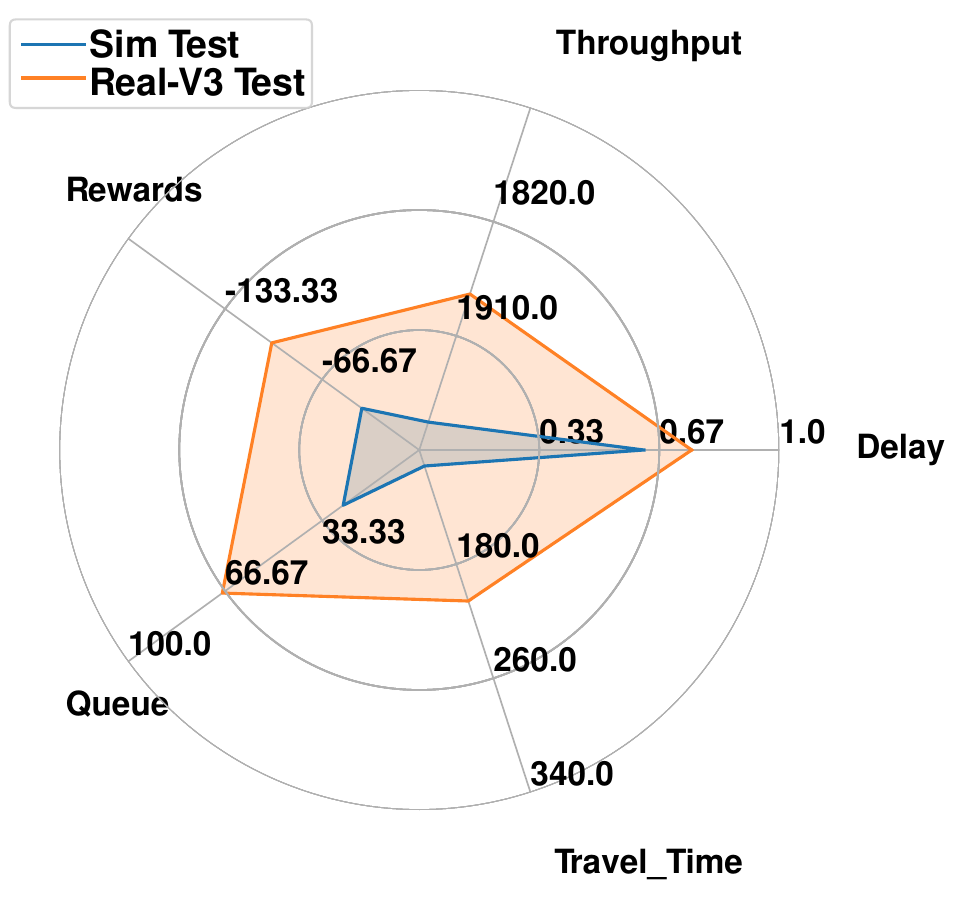}
    &\includegraphics[width=0.220\textwidth]
    {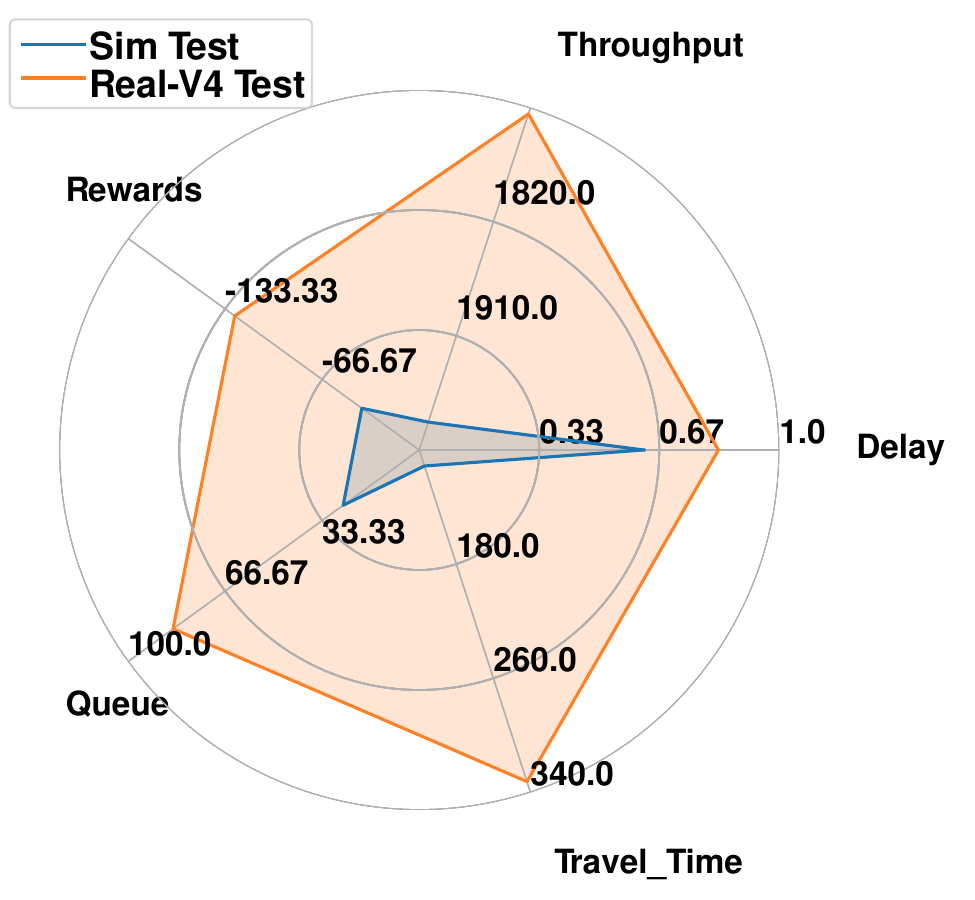}\\

    \end{tabular}
    \caption{The performance gap using Direct-Transfer to train in $E_{sim}$ and tested in 4 $E_{real}$ settings.}
    \label{fig:main}
    \vspace{-3mm}
\end{figure}

To investigate the presence of a performance gap in TSC tasks, we conducted an experiment. We employed the Direct-Transfer method, training a policy model $\pi_{test}$ in $E_{sim}$ using the DQN method for 300 epochs. We then directly transferred $\pi_{test}$ to four distinct $E_{real}$ settings, as detailed in Table~\ref{param}.
The results are visualized in Fig.~\ref{fig:main}, using a radar chart with five metrics indicating performance in two environments, $E_{sim}$ and $E_{real}$. The blue line connects the metrics in $E_{sim}$, while the orange line represents $E_{real}$. A notable performance gap emerges when applying $\pi_{test}$ to four $E_{real}$ settings compared to $E_{sim}$. Our experiment confirms the existence of performance gaps, prompting further study into method generalizability.

\begin{table}[htb]
    \centering
    \caption{Direct-Transfer and \ours performance in $E_{sim}$}
    \scalebox{0.95}{
    \begin{tabular}{cccccc}
        \toprule
        $Env$ & $ATT$ & $TP$ & $Reward$ & $Queue$ & $Delay$\\ \midrule
        \textbf{$E_{sim}$} & \textbf{111.23$_{\pm{\text{0.05}}}$} & \textbf{1978$_{\pm{\text{1}}}$} &\textbf{-39.44$_{\pm{\text{0.03}}}$} &\textbf{26.11$_{\pm{\text{0.05}}}$} &\textbf{0.62$_{\pm{\text{0.01}}}$} \\
        \bottomrule
    \end{tabular}}
    \label{tab:simresult}
    \vspace{-3mm}
\end{table}

\subsubsection{\textbf{Gap mitigating by uncertainty-aware \ours}}

To verify whether the \ours can effectively mitigate the performance gap, we compare the performance of directly transferring policies trained in $E_{sim}$ to $E_{real}$ with the policies learned under \ours in four $E_{real}$ settings. Because they are using the same $E_{sim}$, so performance in $E_{sim}$ eventually converges to stable results with tiny variance as shown in Table~\ref{tab:simresult}.  

In Table~\ref{tab:main}, we use ($\cdot$) to quantify the performance gap from $E_{real}$ to $E_{sim}$, as computed with Equation \eqref{eq:delta}. This gap directly reflects the methods' generalization capability from $E_{sim}$ to $E_{real}$. Our findings are as follows: (1) When comparing \ours to Direct-Transfer, it effectively reduces the performance gap ($\Delta$). Notably, \ours exhibits smaller gaps in $ATT_{\Delta}$, $Queue_{\Delta}$, and $Delay_{\Delta}$, while showing larger gaps in $TP_{\Delta}$ and $Reward_{\Delta}$, indicating effective performance gap mitigation.
(2) In terms of the original traffic signal control metrics, \ours enhances the performance of policy $\pi_{\phi}$. It achieves lower $ATT$ and higher $TP$ than Direct-Transfer.
(3) Table \ref{tab:main} summarizes experiments across four diverse real-world settings, encompassing five metrics. The results underscore \ours' robustness and effectiveness in complex environmental conditions.




\subsubsection{\textbf{Ablation Study}}


In Table~\ref{tab:ablation}, We conducted an ablation study on \ours to assess the impact of its dynamic grounding module, uncertainty quantification module, and action grounding module. We systematically analyzed each module's influence by removing them step by step. When the dynamic grounding module is removed, $\alpha$ is fixed at 0.5. In the third row, when both $\alpha$ and uncertainty are removed, it becomes Vanilla GAT.

\begin{table}[htb]
    \centering
    \caption{Ablation Study of \ours on $V1$}
    \scalebox{0.82}{
    \begin{tabular}{cccccc}
        \toprule
        $Structure$ 
        
        &\begin{tabular}[c]{@{}c@{}}$ATT_{\Delta}$ \\ $(\Delta\downarrow)$\end{tabular} 

         &\begin{tabular}[c]{@{}c@{}}$TP_{\Delta}$ \\ $(\Delta\uparrow)$\end{tabular} 

         &\begin{tabular}[c]{@{}c@{}}$Reward_{\Delta}$ \\ $(\Delta\uparrow)$\end{tabular} 

         &\begin{tabular}[c]{@{}c@{}}$Queue_{\Delta}$ \\ $(\Delta\downarrow)$\end{tabular} 

         &\begin{tabular}[c]{@{}c@{}}$Delay_{\Delta}$ \\ $(\Delta\downarrow)$\end{tabular} 

        \\ \midrule
        \textbf{UGAT} 
        & \textbf{33.49$_{\pm{\text{3.61}}}$} & \textbf{-52$_{\pm{\text{4.58}}}$} & \textbf{-19.94$_{\pm{\text{3.08}}}$} & \textbf{13.47$_{\pm{\text{2.04}}}$} & \textbf{0.05$_{\pm{\text{0.01}}}$} \\
        
         w/o dynamic $\alpha$  
         &   39.12$_{\pm{\text{4.21}}}$
         &  -72$_{\pm{\text{7.61}}}$
         & -25.07$_{\pm{\text{5.71}}}$
         & 16.88$_{\pm{\text{5.11}}}$
         & 0.08$_{\pm{\text{0.01}}}$\\

         w/o $\alpha$, uncertainty 
         &   44.87$_{\pm{\text{4.81}}}$
         &  -73$_{\pm{\text{12.99}}}$
         & -30.59$_{\pm{\text{3.80}}}$
         & 20.50$_{\pm{\text{1.97}}}$
         & 0.09$_{\pm{\text{0.01}}}$\\
         
         w/o Grounding  
         
        &   47.71$_{\pm{\text{6.73}}}$
         &  -77$_{\pm{\text{10.64}}}$
         & -32.11$_{\pm{\text{4.24}}}$
         & 21.60$_{\pm{\text{3.12}}}$
         & 0.11$_{\pm{\text{0.02}}}$\\
        \bottomrule
    \end{tabular}
    }
    \label{tab:ablation}
\end{table}

We conducted an ablation study in Table~\ref{tab:static} to investigate the impact of dynamically adjusted grounding rates $\alpha$ on sim-to-real training. We activated the uncertainty quantification module EDL and manually set $\alpha$ values ranging from 0.2 to 0.8. In this study, if the model's uncertainty output $u$ was less than the static $\alpha$, the action was grounded; otherwise, it was rejected as in Algorithm \ref{algo:UGAT}. We compared these results with \ours, which dynamically adjusts $\alpha$ based on uncertainty. Notably, \ours significantly improved model performance.

\begin{table}[htb]
    \centering
    \caption{Static vs dynamic $\alpha$ on $V1$}
    \scalebox{0.85}{
    \begin{tabular}{cccccc}
        \toprule
        $\alpha$ 
        
        &\begin{tabular}[c]{@{}c@{}}$ATT_{\Delta}$ \\ $(\Delta\downarrow)$\end{tabular} 

         &\begin{tabular}[c]{@{}c@{}}$TP_{\Delta}$ \\ $(\Delta\uparrow)$\end{tabular} 

         &\begin{tabular}[c]{@{}c@{}}$Reward_{\Delta}$ \\ $(\Delta\uparrow)$\end{tabular} 

         &\begin{tabular}[c]{@{}c@{}}$Queue_{\Delta}$ \\ $(\Delta\downarrow)$\end{tabular} 

         &\begin{tabular}[c]{@{}c@{}}$Delay_{\Delta}$ \\ $(\Delta\downarrow)$\end{tabular} 

        \\ \midrule
        \textbf{dynamic} 
        & \textbf{33.49$_{\pm{\text{3.61}}}$} & \textbf{-52$_{\pm{\text{4.58}}}$} & \textbf{-19.94$_{\pm{\text{3.08}}}$} & \textbf{13.47$_{\pm{\text{2.04}}}$} & \textbf{0.05$_{\pm{\text{0.01}}}$} \\
        
         0.2  
         &   68.59$_{\pm{\text{7.14}}}$
         &  -117$_{\pm{\text{12.53}}}$
         & -40.42$_{\pm{\text{3.92}}}$
         & 27.11$_{\pm{\text{4.29}}}$
         & 0.12$_{\pm{\text{0.05}}}$\\

         0.4 
         &   55.87$_{\pm{\text{7.83}}}$
         &  -73$_{\pm{\text{13.01}}}$
         & -30.69$_{\pm{\text{4.54}}}$
         & 20.30$_{\pm{\text{3.28}}}$
         & 0.12$_{\pm{\text{0.01}}}$\\

        0.5
         &   39.12$_{\pm{\text{4.21}}}$
         &  -72$_{\pm{\text{7.61}}}$
         & -25.07$_{\pm{\text{5.71}}}$
         & 16.88$_{\pm{\text{5.11}}}$
         & 0.08$_{\pm{\text{0.01}}}$\\
         
         0.6
         &   47.09$_{\pm{\text{2.79}}}$
         &  -77$_{\pm{\text{4.68}}}$
         & -34.11$_{\pm{\text{3.99}}}$
         & 21.31$_{\pm{\text{2.38}}}$
         & 0.10$_{\pm{\text{0.03}}}$\\

        0.8
         &   48.53$_{\pm{\text{6.70}}}$
         &  -85$_{\pm{\text{9.17}}}$
         & -37.85$_{\pm{\text{6.23}}}$
         & 25.60$_{\pm{\text{2.91}}}$
         & 0.11$_{\pm{\text{0.01}}}$\\

        \bottomrule
    \end{tabular}
    }
    \label{tab:static}
    \vspace{-3mm}
\end{table}





\subsubsection{\textbf{Different Uncertainty Methods in \ours}}
\label{sec:exp:uncertainty}
\begin{figure}[htb]
\vspace{-4.5mm}
    \centering
    \begin{tabular}{cc}
    \includegraphics[width=0.23\textwidth]{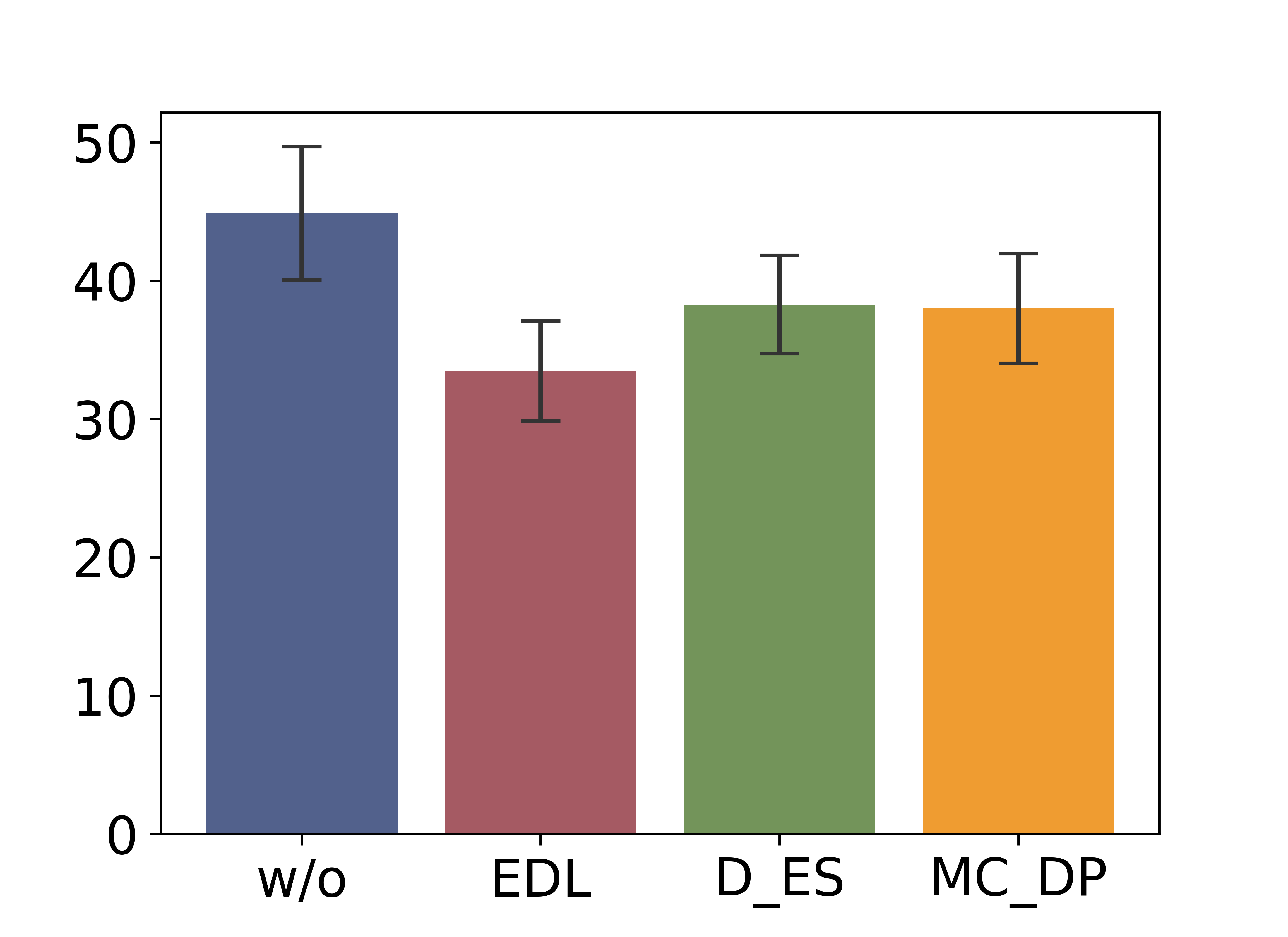} &
    \includegraphics[width=0.23\textwidth]{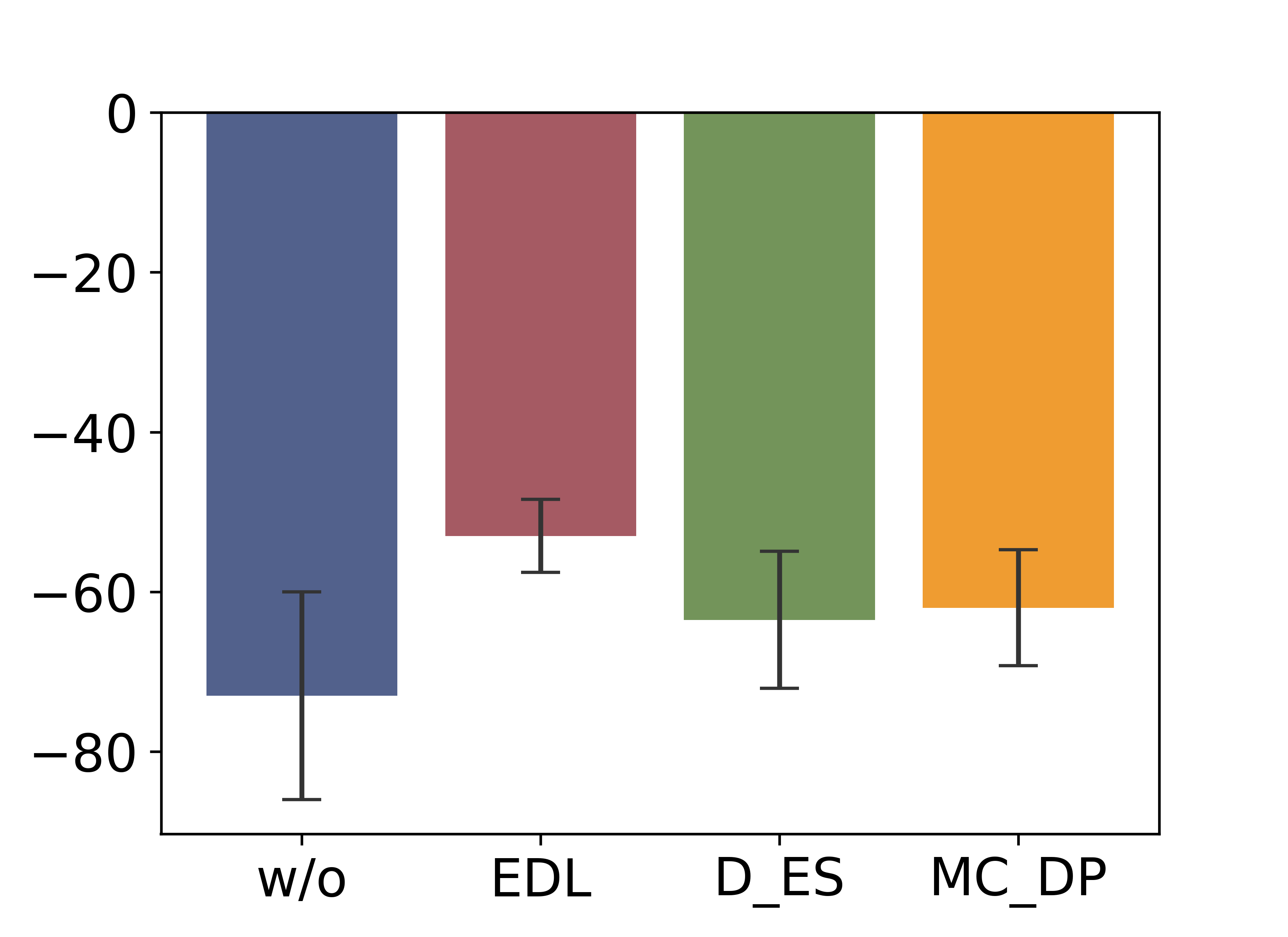} \\
        \small{
        \begin{tabular}[c]{@{}c@{}}(a) $ATT_{\Delta}$ of 4 methods \end{tabular}}& 
        \small{
        \begin{tabular}[c]{@{}c@{}}(b) $TP_{\Delta}$ of 4 methods \end{tabular}}
    \end{tabular}
    \caption{Uncertainty investigation across 4 methods on $V1$}
    \label{fig:uncertainty}
    \vspace{-4mm}
\end{figure}

In our previous experiments, we utilized EDL~\cite{sensoy2018evidential} for uncertainty quantification. To gain deeper insights into the benefits of model uncertainty, we conducted additional experiments using various uncertainty quantification methods, namely, EDL, Concrete Dropout~\cite{gal2017concrete}, and Deep Ensembles~\cite{lakshminarayanan2017simple}. We compared these methods with w/o that excluded the uncertainty module shown in Fig.~\ref{fig:uncertainty}, smaller $ATT_{\Delta}$ and larger $TP_{\Delta}$ is better, which demonstrate model uncertainty quantification methods narrow the performance gap, with EDL outperforming the others, validating its use in \ours.




%

\vspace{-5mm}
\section{Related Work}


\paragraph{Sim-to-real Transfer}
The literature on sim-to-real transfer~\cite{zhao2020sim} can be generally categorized into three groups. The \textbf{\emph{domain randomization}}~\cite{tobin2019real} aims to learn policies that are resilient to changes in the environment. The \textbf{\emph{domain adaptation}}~\cite{tzeng2019deep} tackles the domain distribution shift problem by unifying the source domain and the target domain features that mainly applied in the perception of robots~\cite{james2019sim}, whereas in TSC, the gap is mainly from the dynamics. The \textbf{\emph{grounding methods}}, improve the accuracy of the simulator concerning the real world by correcting simulator bias. 
Grounded Action Transformation~\cite{hanna2017grounded} induces the dynamics of the simulator to match reality-grounded action,~\cite{IROS20-Desai, IROS20-Karnan, NEURIPS20-Karnan} further explore modeling the stochasticity when grounding,  applying RL, and Imitation from observation (IfO) techniques to advance grounding. Inspired by the above work, \ours leverages uncertainty quantification to enhance action transformation.

\paragraph{Uncertainty Quantification}
Effective uncertainty quantification (UQ) is essential in current deep learning methods to grasp model limitations and enhance model acceleration and accuracy. Gaussian Process (GPs) \cite{seeger2004gaussian} is a non-parametric approach for quantifying uncertainty, while another line of research involves using prior distributions on model parameters to estimate uncertainty during training \cite{xue2019reliable}. Evidential Deep Learning (EDL) \cite{sensoy2018evidential}, MC dropout \cite{gal2017concrete}, and Deep Ensembles \cite{lakshminarayanan2017simple} are representative methods that leverage parametric models. This paper experiments on EDL, MC dropout, and Deep Ensembles to explore their benefits.


\vspace{-4.5mm}
\section{Conclusion}
In this paper, we identify the performance gap in traffic signal control problems and introduce \ours, an uncertainty-aware grounding action transformation method, to dynamically adapt actions in simulations. Our experiments confirm that \ours effectively reduces the performance gap with improved stability. This work represents progress in enhancing the real-world applicability of RL-based traffic signal control models, our code can be found at  \url{https://github.com/DaRL-LibSignal/UGAT.git}.
\vspace{-5mm}


\bibliographystyle{IEEEtran}
\vspace{-5mm}
\bibliography{software.bib}

\end{document}